\documentclass[10pt,twocolumn,letterpaper]{article}

\usepackage{iccv}
\usepackage{times}
\usepackage{epsfig}
\usepackage{graphicx}
\usepackage{amsmath}
\usepackage{amssymb}
\usepackage{adjustbox}
\usepackage{caption}
\usepackage{float}
\usepackage{microtype}
\usepackage{subfigure}
\usepackage{enumitem}
\usepackage{makecell}    
\usepackage{pifont}  
\usepackage{booktabs} 
\usepackage{times}
\usepackage{epsfig}
\usepackage{graphicx}
\usepackage{amsmath}
\usepackage{siunitx}
\usepackage{xcolor}
\usepackage{stfloats}
\usepackage[utf8]{inputenc}

\newcommand{\xmark}{\ding{55}} 

\usepackage{color}

\definecolor{darkgreen}{RGB}{0,127,0}
\definecolor{darkred}{RGB}{200,0,0}
\def\greencheckmark{\textcolor{darkgreen}{\checkmark}}
\def\redxmark{\textcolor{darkred}{\xmark}}
\newcommand{\green}[1] {\footnotesize \color[rgb]{0.13, 0.55, 0.13}(\ensuremath #1\%)}
\usepackage{colortbl}

\usepackage{multirow}
\usepackage{tabularx}
\usepackage{hhline}
\usepackage[para,online,flushleft]{threeparttable}
\usepackage{stackengine,scalerel}

\newcommand\overstar[1]{\ThisStyle{\ensurestackMath{%
  \setbox0=\hbox{$\SavedStyle#1$}%
  \stackengine{0mm}{\copy0}{\kern.2\ht0\smash{\SavedStyle*}}{O}{c}{F}{T}{S}}}}

\usepackage[pagebackref=true,breaklinks=true,letterpaper=true,colorlinks,bookmarks=false]{hyperref}

\iccvfinalcopy 

\ificcvfinal\fi

\begin{document}

\title{Distill Any Depth:  \\Distillation Creates a Stronger Monocular Depth Estimator}

\author{Xiankang He$^{*1,2}$ \:\: Dongyan Guo$^{*1}$ \:\: Hongji Li$^{2,3}$ \:\: Ruibo Li$^{4}$ \:\: Ying Cui$^{1}$ \:\: Chi Zhang$^{\dagger2 }$\\
\begin{tabular}[h]{cc}
    $^{1}$Zhejiang University of Technology \quad\quad $^{2}$ AGI Lab, Westlake University\\ 
    $^{3}$Lanzhou University  \quad\quad $^{4}$Nanyang Technological University \\
    {\tt\small \{hexiankang577, 3420670269neon\}@gmail.com \quad \{guodongyan,cuiying\}@zjut.edu.cn} \\
    {\tt\small ruibo001@e.ntu.edu.sg \quad chizhang@westlake.edu.cn} \\
    \url{https://distill-any-depth-official.github.io/}
\end{tabular}
}

\twocolumn[{\maketitle
  \ificcvfinal\thispagestyle{empty}\fi
  \centering
  \vspace{-0.7cm}
  \includegraphics[width=0.97\textwidth]{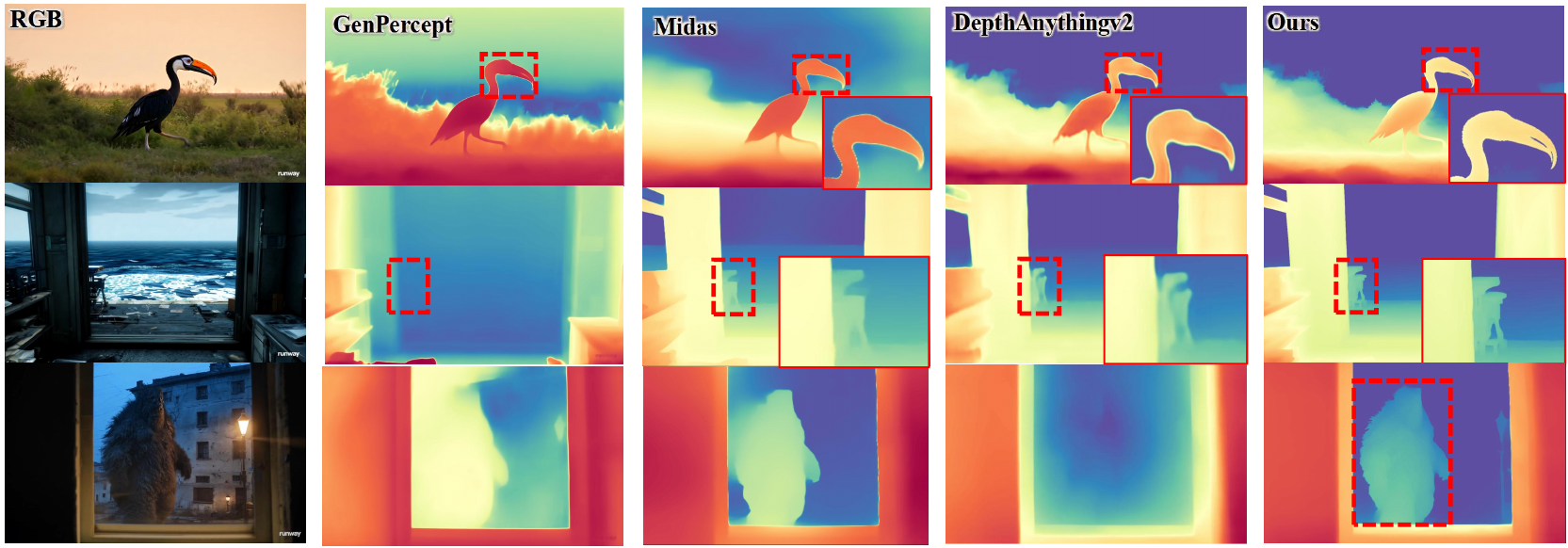}
  \vspace{-.05 in}
  \captionof{figure}{
    \textbf{Zero-shot prediction on in-the-wild images.} Our model, distilled from Genpercept~\cite{xu2024diffusion} and DepthAnythingv2~\cite{depth_anything_v2}, outperforms other methods by delivering more accurate depth details and exhibiting superior generalization for monocular depth estimation on in-the-wild images. 
    }
    \vspace{0.1cm}
    \label{fig:distill_with_mt}
    }
]

{
  \renewcommand{\thefootnote}%
    {\fnsymbol{footnote}}
  \footnotetext[1]{denotes co-first authorship. This work was done while Xiankang He was a visiting student at the AGI Lab, Westlake University.\\
  \hspace*{1.35em}$^{\dagger}$ denotes corresponding author.
  }
}

\begin{abstract}
Recent advances in zero-shot monocular depth estimation(MDE) have significantly improved generalization by unifying depth distributions through normalized depth representations and by leveraging large-scale unlabeled data via pseudo-label distillation. However, existing methods that rely on global depth normalization treat all depth values equally, which can amplify noise in pseudo-labels and reduce distillation effectiveness.
In this paper, we present a systematic analysis of depth normalization strategies in the context of pseudo-label distillation. Our study shows that, under recent distillation paradigms (e.g., shared-context distillation), normalization is not always necessary—omitting it can help mitigate the impact of noisy supervision.
Furthermore, rather than focusing solely on how depth information is represented, we propose Cross-Context Distillation, which integrates both global and local depth cues to enhance pseudo-label quality. We also introduce an assistant-guided distillation strategy that incorporates complementary depth priors from a diffusion-based teacher model, enhancing supervision diversity and robustness.
Extensive experiments on benchmark datasets demonstrate that our approach significantly outperforms state-of-the-art methods, both quantitatively and qualitatively.
\end{abstract}

\section{Introduction}
\label{sec:intro}
Monocular depth estimation (MDE) predicts scene depth from a single RGB image, offering greater flexibility compared to stereo or multi-view methods, and benefiting a wide range of applications, such as autonomous driving and robotic perception~\cite{eigen2014depth,garg2016unsupervised,guizilini20203d,yang2020d3vo,li2020ar}. Recent research on zero-shot MDE models~\cite{ranftl2020midas,yin2020diversedepth,wei2021leres, marigold} aims to handle diverse scenarios, but training such models requires large-scale, diverse depth data, which is often limited by the need for specialized equipment~\cite{mayer2016large,yin2021learning}. A promising solution is using large-scale unlabeled data, which has shown success in tasks like classification and segmentation~\cite{kirillov2023segment,zoph2020rethinking,xie2020self}. Studies like DepthAnything~\cite{yang2024depthanything} highlight the effectiveness of using pseudo labels from teacher models for training student models.

To enable training on such a diverse, mixed dataset, most
state-of-the-art methods~\cite{depth_anything_v2, ranftl2020towards, yin2020diversedepth} employ scale-and-shift invariant (SSI) depth representations for loss computation. This approach normalizes raw depth values within an image, making them invariant to scaling and shifting, and ensures that the model learns to focus on relative depth relationships rather than absolute values. The SSI representation facilitates the joint use of diverse depth data, thereby improving the model's ability to generalize across different scenes~\cite{ranftl2021vision, caron2021emerging}. Similarly, during evaluation, the metric depth of the prediction is recovered by solving for the unknown scale and shift coefficients of the predicted depth using least squares, ensuring the application of standard evaluation metrics.

\begin{figure}[t]
    \centering  \includegraphics[width=0.47\textwidth]{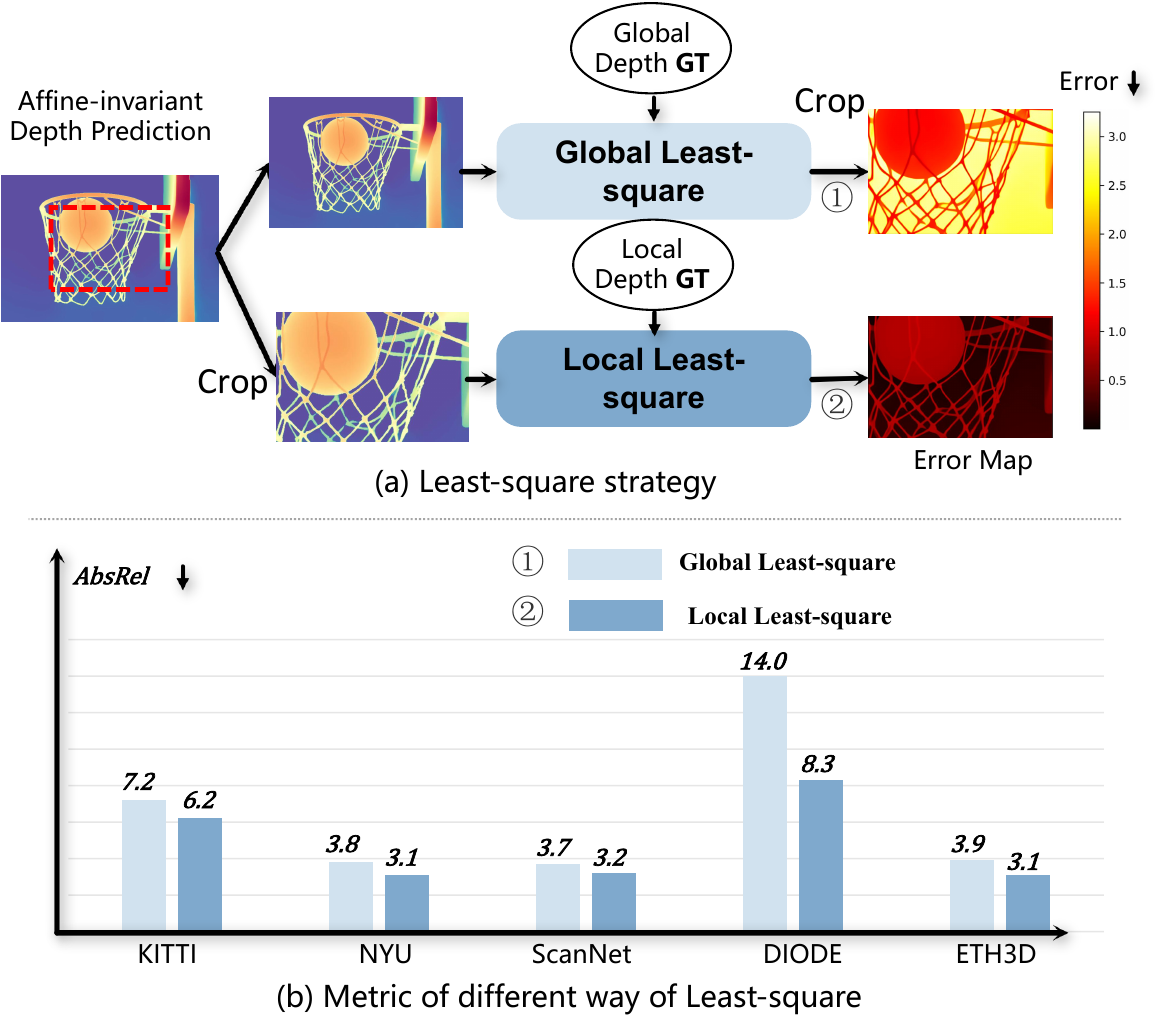}
    \caption{
    \textbf{Issue with Global Normalization (SSI).} 
    In (a), we compare two alignment strategies for the central \( w/2, h/2 \) region: (1) \textit{Global Least-Square}, where alignment is applied to the full image before cropping, and (2) \textit{Local Least-Square}, where alignment is performed on the cropped region. Metrics are computed on the cropped region. As shown in (b), the outperformed local strategy demonstrates
    that \textbf{global normalization degrades local accuracy compared to local normalization}.
    }
    \label{fig:normalization}
\end{figure}

Despite its advantages, using SSI depth representation for pseudo-label distillation in MDE models presents several issues. Specifically, the inherent normalization process in SSI loss makes the depth prediction at a given pixel not only dependent on the teacher model’s raw prediction at that location but also influenced by the depth values in other regions of the image. 
This becomes problematic because pseudo-labels inherently introduce noise. Even if certain local regions are predicted accurately, inaccuracies in other regions can negatively affect depth estimates after global normalization, leading to suboptimal distillation results.
As shown in Fig.~\ref{fig:normalization}, we empirically demonstrate that normalizing depth maps globally tends to degrade the accuracy of local regions, as compared to only applying normalization within localized regions during evaluation.

Building on this insight, in this paper, we investigate the issue of depth normalization in pseudo-label distillation. We analyze various depth normalization strategies, including global normalization, local normalization, hybrid global-local approaches, and the absence of normalization. Through empirical experiments, we explore how each depth representation impacts the performance of different distillation designs, especially when using pseudo-labels for training.

Rather than focusing solely on pseudo-label representation, we introduce Cross-Context Distillation, a method that integrates both global and local depth cues to enhance pseudo-label quality. Our findings reveal that local regions, when used for distillation, produce pseudo-labels that capture higher-quality depth details, improving the student model’s depth estimation accuracy. However, relying solely on local regions may overlook broader contextual relationships in the image. To address this, we combine both local and global inputs within a unified distillation framework. By leveraging the context-specific advantages of local distillation alongside the broader understanding provided by global methods, our approach yields more detailed and reliable depth predictions.

To harness the strengths of both, we propose using a diffusion-based model as the teacher assistant to generate pseudo-labels, which are then used to supervise the student model. This strategy enables the student model to learn from the detailed depth information provided by diffusion-based models, while also benefiting from the precision and efficiency of encoder-decoder models.

To validate the effectiveness of our design, we conduct extensive experiments on various benchmark datasets. The empirical results show that our method significantly outperforms existing baselines qualitatively and quantitatively. The contributions can be summarized below:
1) 
We systematically analyze the role of different depth normalization strategies in pseudo-label distillation, providing insights into their effects on MDE performance.
2) 
To enhance the quality of pseudo-labels, we propose Cross-Context Distillation, a hybrid local-global framework that leverages fine-grained details and global depth relationships; a teacher assistant that harnesses the complementary strengths of diverse depth estimation models to further improve robustness and accuracy.
3) 
We conduct extensive experiments on benchmark datasets, demonstrating that our method outperforms state-of-the-art approaches both quantitatively and qualitatively.

\section{Related Work}

\label{sec:Related}
\subsection{Monocular Depth Estimation}
Monocular depth estimation (MDE) has evolved from hand-crafted methods to deep learning, significantly improving accuracy \cite{eigen2014depth, laina2016deeper, fu2018deep, godard2017unsupervised, zhou2017unsupervised, ranftl2021vision}. Architectural refinements, such as multi-scale designs and attention mechanisms, have further enhanced feature extraction \cite{hu2018squeeze, chen2017deeplab, zhao2017pyramid}. However, most models remain reliant on labeled data and struggle to generalize across diverse environments.
Zero-shot MDE improves generalization by leveraging large-scale datasets, geometric constraints, and multi-task learning \cite{ranftl2020midas,yin2020diversedepth,yin2020learning,zhang2023robust,ge2024geobench}. Metric depth estimation incorporates intrinsic data for absolute depth learning \cite{bhat2023zoedepth, yin2023metric3d, hu2024metric3d,piccinelli2024unidepth, wang2024moge, depthpro}, while generative models such as Marigold refine depth details using diffusion priors \cite{marigold, xu2024diffusion, gui2024depthfm, he2024lotus}. 
Despite these advances, effectively utilizing unlabeled data remains a challenge due to pseudo-label noise and inconsistencies across different contexts. DepthAnything \cite{depth_anything_v2} explores large-scale unlabeled data but struggles with pseudo-label reliability. PatchFusion \cite{patchfusion2023, miangoleh2021boosting} improves depth estimation by refining high-resolution image representations but lacks adaptability in generative settings.
To address these issues, we propose Cross-Context and Multi-Teacher Distillation, which enhances pseudo-label supervision by leveraging diverse contextual information and multiple expert models, improving both accuracy and generalization ability.

\subsection{Semi-supervised Monocular Depth Estimation}

Semi-supervised depth estimation has garnered increasing attention, primarily leveraging temporal consistency to utilize unlabeled data more effectively~\cite{kuznietsov2017semi, guizilini2020robust}. Some approaches~\cite{left_right, smolyanskiy2018importance, cho2019large, yang2018deep, monodepth2} integrate stereo geometric constraints, enforcing left-right consistency in stereo video to enhance depth accuracy. Others incorporate additional supervision, such as semantic priors~\cite{ramirez2018geometry,hoyer2023improving}or generative adversarial networks (GANs). For instance, DepthGAN~\cite{gandepth} refines depth predictions through adversarial learning. However, these methods often rely on temporal cues, stereo constraints, or other auxiliary information, limiting their applicability to broader and more general scenarios. Recent work~\cite{petrovai2022exploiting} has explored pseudo-labeling for semi-supervised monocular depth estimation (MDE), but it lacks generative modeling capabilities, restricting its generalization across diverse environments. DepthAnything ~\cite{yang2024depthanything} demonstrates the effectiveness of large-scale unlabeled data in improving generalization; however, pseudo-label reliability remains a challenge. In contrast, our approach focuses on single-image depth estimation, improving pseudo-label reliability and maximizing its effectiveness. By relying solely on unlabeled data without additional constraints, our method achieves a more accurate and generalizable MDE model.

\begin{figure*}[!t]
    \centering    \includegraphics[width=1\textwidth]{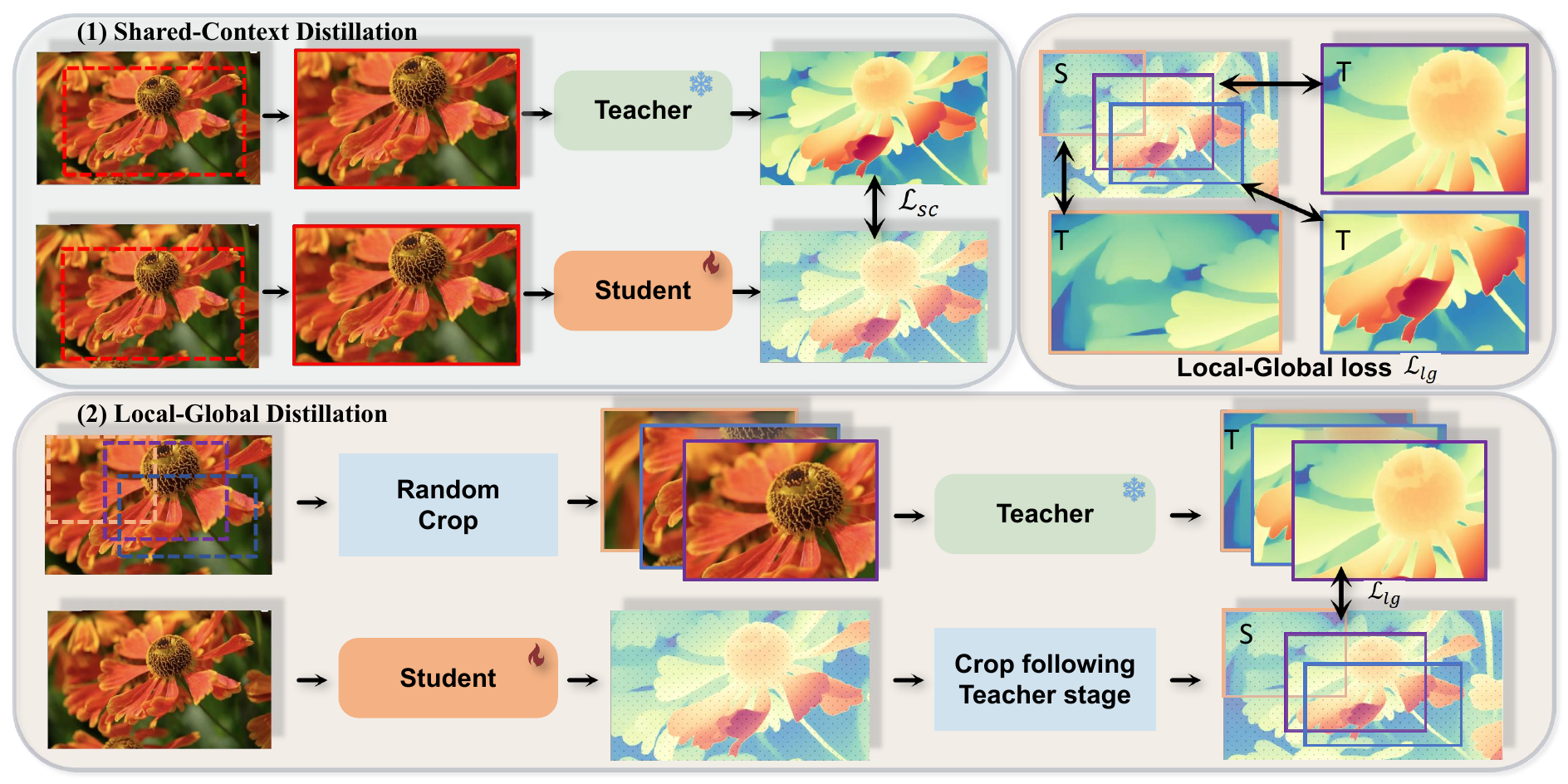}
    \caption{
    \textbf{Overview of Cross-Context Distillation.} 
    Our method combines local and global depth information to enhance the student model’s predictions. It includes two scenarios: (1) \textit{Shared-Context Distillation}, where both models use the same image for distillation;
    and (2) \textit{Local-Global Distillation}, where the teacher predicts depth for overlapping patches while the student predicts the full image. The Local-Global loss $\mathcal{L}_{\text{lg}}$ (Top Right) ensures consistency between local and global predictions, enabling the student to learn both fine details and broad structures, improving accuracy and robustness.
    }
    \label{fig:method}
\end{figure*}

\section{Method}
\label{sec:Method}
In this section, we introduce a novel distillation framework designed to leverage unlabeled images for training zero-shot Monocular Depth Estimation (MDE) models. We begin by exploring various depth normalization techniques in Section~\ref{sec:depth_norm}, followed by detailing our proposed distillation method in Section~\ref{sec:ref_dis}, which combines predictions across multiple contexts. The overall framework is illustrated in Fig.~\ref{fig:method}. Finally, we introduce an assistant-guided distillation scheme, in which a diffusion-based model acts as an auxiliary teacher to provide additional supervision for student training.

\subsection{Depth Normalization}
\label{sec:depth_norm}
\begin{figure}[t]  
    \centering  
      \includegraphics[width=0.5\textwidth]{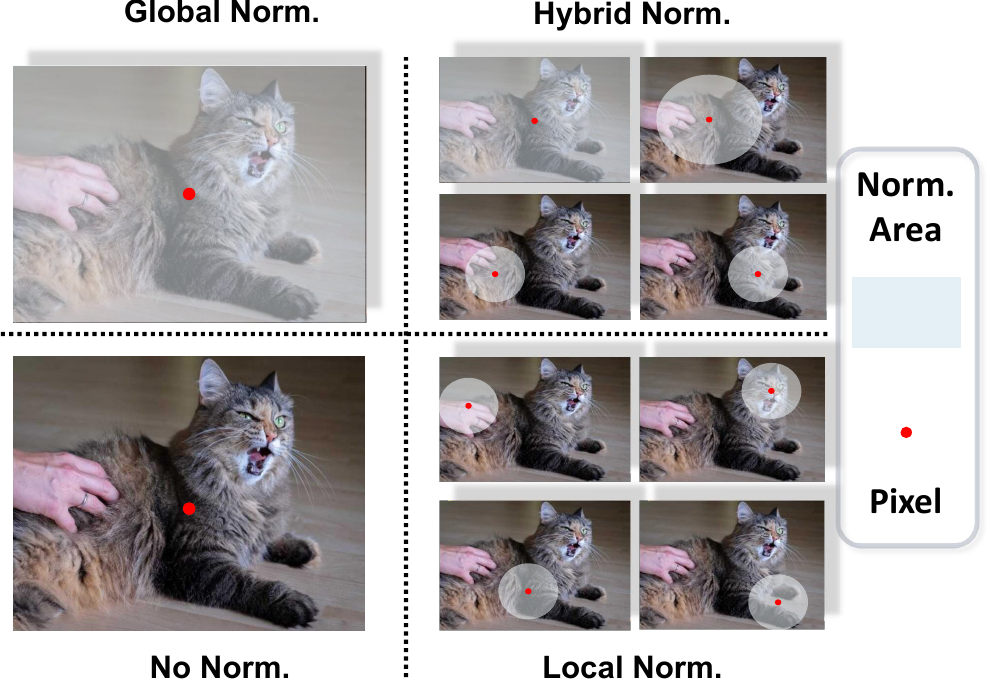}  
    \caption{\textbf{Normalization Strategies.} We compare four normalization strategies: Global Norm~\cite{ranftl2020midas}, Hybrid Norm~\cite{zhang2022hdn}, Local Norm, and No Norm. The figure visualizes how each strategy processes pixels within the normalization region (Norm. Area). The red dot represents any pixel within the region.
    }

    \label{fig:norm}
\end{figure}

Depth normalization is a crucial component of our framework as it adjusts the pseudo-depth labels \( \mathbf{d}^t \) from the teacher model and the depth predictions \( \mathbf{d}^s \) from the student model for effective loss computation. To understand the influence of normalization techniques on distillation performance, we systematically analyze several approaches commonly employed in prior works. These strategies are visually illustrated in Fig.~\ref{fig:norm}.

\noindent \textbf{Global Normalization:} The first strategy we examine is the global normalization ~\cite{yang2024depthanything,depth_anything_v2} used in recent distillation methods. 
Global normalization~\cite{ranftl2020midas} adjusts depth predictions using global statistics of the entire depth map. This strategy aims to ensure scale-and-shift invariance by normalizing depth values based on the median and mean absolute deviation of the depth map. For each pixel \( i \), the normalized depth for the student model and pseudo-labels are computed as:
\begin{equation}
\begin{aligned}
\tilde{d}^s_i &= \mathcal{N}_{glo}(\mathbf{d}^s) = \frac{d^s_{i} - \operatorname{med}(\mathbf{d}^s)}{\frac{1}{M} \sum_{j=1}^M \left| d^s_{j} - \operatorname{med}(\mathbf{d}^s) \right|} \\
\tilde{d}^t_i &= \mathcal{N}_{glo}(\mathbf{d}^t) = \frac{d^t_{i} - \operatorname{med}(\mathbf{d}^t)}{\frac{1}{M} \sum_{j=1}^M \left| d^t_{j} - \operatorname{med}(\mathbf{d}^t) \right|},
\end{aligned}
\end{equation}
where \( \operatorname{med}(\mathbf{d}^s) \) and \( \operatorname{med}(\mathbf{d}^t) \) are the medians of the predicted depth and pseudo depth, respectively. The final regression loss for distillation is computed as the average absolute difference between the normalized predicted depth and the normalized pseudo depth across all valid pixels \( M \):
\begin{equation}
\mathcal{L}_{\text{Dis}} = \frac{1}{M} \sum_{i=1}^M \left| \tilde{d}^s_i - \tilde{d}^t_i \right|.
\label{eq:ssi}
\end{equation}

\noindent \textbf{Hybrid Normalization:}
In contrast to global normalization, Hierarchical Depth Normalization~\cite{zhang2022hdn} employs a hybrid normalization approach by integrating both global and local depth information. This strategy is designed to preserve both the global structure and local geometry in the depth map.
The process begins by dividing the depth range into \( S \) segments, where \( S \) is selected from \( \{ 1, 2, 4 \} \). When \( S = 1 \), the entire depth range is normalized globally, treating all pixels as part of a single context, akin to global normalization.
In the case of \( S = 2 \), the depth range is divided into two segments, with each pixel being normalized within one of these two local contexts. Similarly, for \( S = 4 \), the depth range is split into four segments, allowing normalization to be performed within smaller, localized contexts.
By adapting the normalization process to multiple levels of granularity, hybrid normalization achieves a balance between global coherence and local adaptability.
For each context \( u \), the normalized depth values for the student model \( \mathcal{N}_u(d_i^s) \) and pseudo-labels \( \mathcal{N}_u(d_i^t) \) are calculated within the corresponding depth range. The loss for each pixel \( i \) is then computed by averaging the losses across all contexts \( U_i \) to which the pixel belongs:
\begin{equation}
\mathcal{L}_{Dis}^i = \frac{1}{|U_i|} \sum_{u \in U_i} \left| \mathcal{N}_u(d_i^s) - \mathcal{N}_u(d_i^t) \right|,
    \label{eq:hdn}
\end{equation}
where \( |U_i| \) denotes the total number of groups (or contexts) that pixel \( i \) is associated with.
To obtain the final loss \( \mathcal{L}_{\text{Dis}} \), we average the pixel-wise losses across all valid pixels \( M \):
\begin{equation}
    \mathcal{L}_{\text{Dis}} = \frac{1}{M} \sum_{i=1}^M \mathcal{L}_{Dis}^i.
\end{equation}

\noindent \textbf{Local Normalization:} In addition to global and hybrid normalization, we investigate Local Normalization, a strategy that focuses exclusively on the finest-scale groups used in hybrid normalization. This approach isolates the smallest local contexts for normalization, emphasizing the preservation of fine-grained depth details without considering hierarchical or global scales.
Local normalization operates by dividing the depth range into the smallest groups, corresponding to \( S = 4 \) in the hybrid normalization framework, and each pixel is normalized within its local context.
The loss for each pixel \( i \) is computed using a similar formulation as in hybrid normalization, but with \( u^i \) now representing the local context for pixel \( i \), defined by the smallest four-part group:
\begin{equation}
    \mathcal{L}_{\text{Dis}} = \frac{1}{M} \sum_{i=1}^M \left| \mathcal{N}_{u^i}(d_i^s) - \mathcal{N}_{u^i}(d_i^t) \right|.
\end{equation}

\noindent\textbf{No Normalization:}
As a baseline, we also consider a direct depth regression approach with no explicit normalization. The absolute difference between raw student predictions and teacher pseudo-labels is used for loss computation:
\begin{equation}
\mathcal{L}_{\text{Dis}} = \frac{1}{M} \sum_{i=1}^M \left| d^s_i - d^t_i \right|,
\label{eq:l1_loss}
\end{equation}

This approach eliminates the need for normalization, assuming pseudo-depth labels naturally reside in the same domain as predictions. It provides insight into whether normalization enhances distillation effectiveness or if raw depth supervision suffices.

\subsection{Distillation Pipeline}
\label{sec:ref_dis}
In this section, we introduce an enhanced distillation pipeline that integrates two complementary strategies: Cross-Context Distillation andassistant-guided distillation. Both strategies aim to improve the quality of pseudo-label distillation, enhance the model’s fine-grained perception.

\noindent \textbf{Cross-context Distillation.}  
A key challenge in monocular depth distillation is the trade-off between local detail preservation and global depth consistency. As shown in Fig.~\ref{fig:context_dis}, providing a local crop of an image as input to the teacher model enhances fine-grained details in the pseudo-depth labels, but it may fail to capture the overall scene structure. Conversely, using the entire image as input preserves the global depth structure but often lacks fine details. To address this limitation, we propose Cross-Context Distillation, a method that enables the student model to learn both local details and global structures simultaneously. Cross-context distillation consists of two key strategies:

\begin{figure}[t]
    \centering  
    \includegraphics[width=0.45\textwidth]{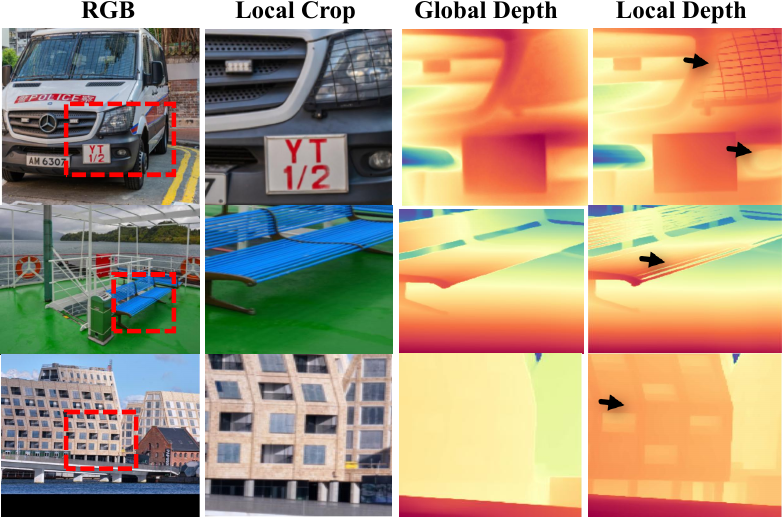}  
    \caption{\textbf{Different Inputs Lead to Different Pseudo Labels.} Global Depth: The teacher model predicts depth using the entire image, and the local region's prediction is cropped from the output. Local Depth: The teacher model directly takes the cropped local region as input, resulting in more refined and detailed depth estimates for that area, capturing finer details compared to using the entire image.}

    \label{fig:context_dis}
\end{figure}

\noindent \textbf{1) Shared-Context Distillation:} In this setup, both the teacher and student models receive the same cropped region of the image as input. Instead of using the full image, we randomly sample a local patch of varying sizes from the original image and provide it as input to both models. This encourages the student model to learn from the teacher model across different spatial contexts, improving its ability to generalize to varying scene structures.
For the loss of shared-context distillation, the teacher and student models receive identical inputs and produce each depth prediction, denoted as \( \mathbf{d}^t_{\text{local}} \) and \( \mathbf{d}^s_{\text{local}} \):
\begin{equation}
\mathcal{L}_{\text{sc}} = \mathcal{L}_{\text{Dis}}\left (\mathbf{d}^s_{\text{local}}, \mathbf{d}^t_{\text{local}} \right),
\end{equation}
This loss encourages the student model to refine its fine-grained predictions by directly aligning with the teacher’s outputs at local scales.

\noindent \textbf{2) Local-Global Distillation:} In this approach, the teacher and student models operate on different input contexts. The teacher model processes local cropped regions, generating fine-grained depth predictions, while the student model predicts a global depth map from the entire image. To ensure knowledge transfer, the teacher’s local depth predictions supervise the corresponding overlapping regions in the student’s global depth map. This strategy allows the student to integrate fine-grained local details into its holistic depth estimation.
Formally, the teacher model produces multiple depth predictions for cropped regions, denoted as \( \mathbf{d}^t_{\text{local}_n} \), while the student generates a global depth map, \( \mathbf{d}^s_{\text{global}} \). The loss for  Local-Global distillation is computed only over overlapping areas between the teacher's local predictions and the corresponding regions in the student’s global depth map:
\begin{equation}
\mathcal{L}_{\text{lg}} = \frac{1}{N} \sum_{n=1}^N \mathcal{L}_{\text{Dis}}\left( \text{Crop}(\mathbf{d}^s_{\text{global}}),  \mathbf{d}^t_{\text{local}_n} \right),
\end{equation}
where \( \text{Crop}(\cdot) \) extracts the overlapping region from the student’s depth prediction, and \( N \) is the total number of sampled patches. This loss ensures that the student benefits from the detailed local supervision of the teacher model while maintaining global depth consistency.
The total loss function integrates both local and cross-context losses along with additional constraints, including feature alignment and gradient preservation, as proposed in prior works~\cite{depth_anything_v2}:
\begin{equation}
\mathcal{L}_{\text{total}} = \mathcal{L}_{\text{sc}} + \lambda_1 \cdot \mathcal{L}_{\text{lg}} + \lambda_2 \cdot \mathcal{L}_{\text{feat}} + \lambda_3 \cdot \mathcal{L}_{\text{grad}}.
\end{equation}
Here, \( \lambda_1 \), \( \lambda_2 \), and \( \lambda_3 \) are weighting factors that balance the different loss components. 
By incorporating cross-context supervision, this framework effectively allows the student model to integrate both fine-grained details from local crops and structural coherence from global depth maps.



\noindent \textbf{Assistant-Guided Distillation.}
In addition to cross-context distillation, we propose an assistant-guided distillation strategy to further enhance the quality and robustness of the distilled depth knowledge. This approach pairs a primary teacher~\cite{depth_anything_v2} with a single auxiliary assistant, selected as a diffusion-based depth estimator~\cite{xu2024diffusion}, which leverages generative priors to complement the primary teacher’s predictions.
This design leverages their complementary strengths: the primary teacher excels in providing efficient and globally consistent supervision, while the assistant offers fine-grained depth cues derived from its generative modeling capabilities.
By drawing supervision from two distinct architectures trained with different paradigms(e.g., optimization strategies or data distributions), the student benefits from diverse knowledge sources, effectively mitigating biases and limitations inherent to a single teacher model.
Formally, let \( \mathcal{M} \) and \( \mathcal{M}_a \) denote the primary and assistant models, respectively. During training, a probabilistic sampling mechanism determines whether supervision for each iteration is drawn from \( \mathcal{M} \) or \( \mathcal{M}_a \). This stochastic guidance encourages the student to adapt to multiple supervision styles, fostering richer and more generalizable depth representations.
Overall, this assistant-guided scheme introduces complementary and diversified pseudo-labels, reducing over-reliance on any single teacher and improving both generalization and depth estimation performance.


\section{Experiment}
\subsection{Experimental Settings}

\noindent \textbf{Datasets.} 
We train our proposed distillation framework on a subset of 200,000 unlabeled images from the SA-1B dataset~\cite{sa1b}, following the training protocol of DepthAnythingv2~\cite{depth_anything_v2}.
For evaluation, we assess the distilled student model on five widely used depth estimation benchmarks. All test datasets are kept unseen during training, enabling a zero-shot evaluation of generalization performance. The chosen benchmarks include:
NYUv2~\cite{silberman2012indoor}, KITTI~\cite{geiger2012we}, ETH3D~\cite{schoeps2017eth3d}, ScanNet~\cite{dai2017scannet}, and DIODE~\cite{vasiljevic2019diode}. Additional dataset details are provided in the Appendix.

\noindent \textbf{Metrics.} We assess depth estimation performance using two key metrics: the mean absolute relative error (AbsRel) and $\delta_1$ accuracy. 
Following previous studies~\cite{ranftl2020midas,yin2023metric3d, marigold} on zero-shot MDE, we align predictions with ground truth in both scale and shift before evaluation.

\noindent \textbf{Implementation.}
Our experiments use state-of-the-art monocular depth estimation models as teachers to generate pseudo-labels, supervising various student models in a distillation framework with only RGB images as input. 
In shared-context distillation, both teacher and student receive the same global region, extracted via random cropping from the original image. The crop maintains a 1:1 aspect ratio and is sampled within a range from 644 pixels to the shortest side of the image, then resized to \( 560 \times 560 \) for prediction.
In global-local distillation, the global region is cropped into overlapping local patches, each sized \( 560 \times 560 \), for the teacher model to predict pseudo-labels. For assistant-guided distillation, we adopt a probabilistic sampling strategy where the primary teacher and the assistant model are selected with a ratio of 7:3, respectively.
We train our best student model using the distillation pipeline for 20,000 iterations with a batch size of 8 on a single NVIDIA V100 GPU, initialized with pre-trained DAv2-Large weights.
The learning rate is in tune with that of the corresponding student model. For DAv2~\cite{depth_anything_v2}, the decoder learning rate is set to \( 5 \times 10^{-5} \). For the total loss function, we set the parameters as follows: \( \lambda_1 = 0.5 \), \( \lambda_2 = 1.0 \) and \( \lambda_3 = 2.0 \).

\subsection{Analysis}
For the ablation study and analysis, we sample a subset of 50K images from \textbf{SA-1B}~\cite{sa1b} as our training data, with an input image size of 560 × 560 for the network. We conduct experiments on two of the most challenging benchmarks, DIODE~\cite{vasiljevic2019diode} and ETH3D~\cite{schoeps2017eth3d}, which include both indoor and outdoor scenes. 

\noindent \textbf{Analysis of Normalization across Cross-Context Distillation.}
We evaluate the impact of different depth normalization strategies on Cross-Context Distillation, as shown in Table~\ref{impact_of_norm}. The results reveal that the optimal normalization method depends on the specific distillation design.
In shared-context distillation, where all pseudo-labels are generated by a single teacher model, hybrid normalization achieves the best performance, closely followed by no normalization. The consistent domain across supervision signals reduces the need for normalization, enabling the model to better preserve local depth relationships. Unlike ground-truth-based training—where normalization is essential to align depth distributions across datasets captured by heterogeneous sensors or represented in varying formats (e.g., sparse vs. dense, relative vs. absolute)—pseudo-labels from a single model are inherently more uniform. Therefore, direct L1 loss without normalization can more faithfully supervise pixel-level depth without distortion from global rescaling. In contrast, global normalization introduces undesirable inter-pixel dependencies, while local normalization discards global structural coherence.
In local-global distillation, hybrid normalization again proves most effective, likely due to its hierarchical design that enforces consistency across both local and global depth predictions. The relatively small gap between hybrid and global normalization suggests that our framework, which uses local cues to refine global predictions, effectively mitigates the limitations of global normalization. However, no normalization leads to a notable performance drop compared to the shared-context setting, indicating that localized regions in this case come from distinct depth domains, making direct L1 supervision less reliable. Local normalization, as before, sacrifices global consistency and thus underperforms.

\begin{figure*}[t]
    \centering
    \includegraphics[width=1\textwidth]{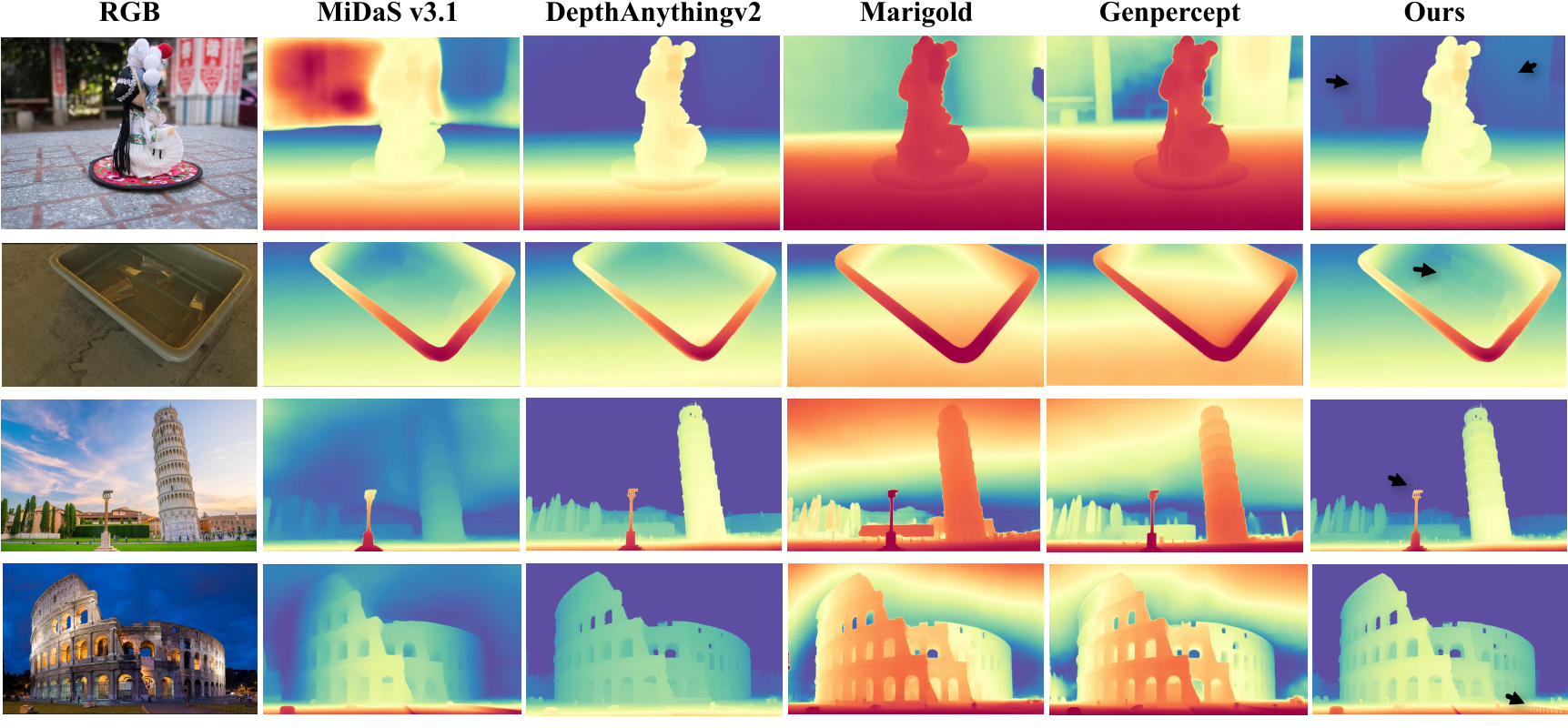}
    \vspace{-2em}
    \caption{
    \textbf{Qualitative Comparison of Relative Depth Estimations.} We present visual comparisons of depth predictions from our method ("Ours") alongside other classic depth estimators (''MiDaS v3.1''~\cite{birkl2023midasv31model}, and models using DINOv2~\cite{oquab2023dinov2} or SD as priors (''DepthAnythingv2~\cite{depth_anything_v2}'', ''Marigold''~\cite{marigold}, ''Genpercept''~\cite{xu2024diffusion}). Compared to state-of-the-art methods, the depth map produced by our model, particularly at the position indicated by the \textbf{black arrow}, exhibits finer granularity and more detailed depth estimation. 
    }

    \label{fig:vis}
\end{figure*}

\begin{table}[t]
    \centering
    \caption{\textbf{Analysis of Normalization Strategies.} Performance comparison of different normalization strategies across Shared-Context Distillation and Local-Global Distillation.}
    \label{impact_of_norm}
    \resizebox{0.5\textwidth}{!}{
        \begin{tabular}{c c S[table-format=1.3] S[table-format=1.3]}
        \toprule
        \multirow{2}{*}{Method} &
        \multirow{2}{*}{Normalization} & 
        \multicolumn{1}{c}{ETH3D} & \multicolumn{1}{c}{DIODE} \\
        && \multicolumn{1}{c}{AbsRel↓} & \multicolumn{1}{c}{AbsRel↓} \\
        \midrule
        \multirow{4}{*}{\shortstack{Shared-Context \\ Distillation}} 
        & Global Norm. & 
        0.064 & 0.259 \\
        \cmidrule{2-4}
        & No Norm. & 
        \textbf{0.057} & 0.239 \\
        & Local Norm. & 
        0.070 & 0.245 \\
        & Hybrid Norm. & 
        \textbf{0.057} & \textbf{0.238} \\
        \midrule
        \multirow{4}{*}{\shortstack{Local-Global \\ Distillation}} 
        & Global Norm. & 
        0.065 & 0.239 \\
        & No Norm. & 
        0.273 & 0.300 \\
        & Local Norm. & 
        0.076 & 0.244 \\
        & Hybrid Norm. & 
        \textbf{0.064} & \textbf{0.238} \\
        \bottomrule
        \end{tabular}
    }
    
\end{table}

\begin{table}[t]
    \caption{\textbf{Effect of Cross-context Distillation.} Performance comparison of various combinations of Shared-Context Distillation and Local-Global distillation on the ETH3D~\cite{schoeps2017eth3d} and DIODE~\cite{vasiljevic2019diode} datasets. The baseline corresponds to a simple shared-context approach with no random cropping. When neither method is applied, the model defaults to this baseline.}
    \label{tab:performance_comparison}
    \resizebox{0.5\textwidth}{!}{ 
    \begin{tabular}{ccll}
    \toprule
    \multirow{2}{*}{\makecell{Shared-Context \\ Distillation}} & \multirow{2}{*}{\makecell{Local-Global \\ Distillation}} & \multicolumn{1}{c}{ETH3D} & \multicolumn{1}{c}{DIODE} \\
    & & \multicolumn{1}{c}{AbsRel↓} & \multicolumn{1}{c}{AbsRel↓} \\
    \midrule
    \redxmark & \redxmark & 0.075 & 0.270 \\
    \redxmark & \greencheckmark & 0.064 \green{-14.6} & 0.238 \green{-13.3} \\
    \greencheckmark & \redxmark & 0.058 \green{-22.6} & 0.237  \green{-12.2}\\
    \rowcolor[rgb]{ .886,  .937,  .855} \greencheckmark & \greencheckmark & \textbf{0.056} \green{-25.3} & \textbf{0.232}  \green{-14.1}\\
    \bottomrule
    \end{tabular}
    }
        
\end{table}

\begin{table}[t]    
    \caption{\textbf{Comparison in Cross-Architecture Distillation.} Evaluation of our distillation pipeline in the context of Cross-Architecture Distillation. We adopt different architectures as teacher and student models, where the \textbf{Base} represents the previous distillation method~\cite{depth_anything_v2}. Our method consistently improves the performance of the distilled student models.}
    \label{tab:cross_architecture_distillation}
    
    \resizebox{0.5\textwidth}{!}{ 
    \begin{tabular}{c c c >{\raggedright\arraybackslash}p{2.2cm} 
    >{\raggedright\arraybackslash}p{2.2cm} }
    \toprule
    \multirow{2}{*}{Teacher} & 
    \multirow{2}{*}{Student} & 
    \multirow{2}{*}{\makecell{Training \\ Loss}} & 
    \multicolumn{1}{c}{DIODE} & 
    \multicolumn{1}{c}{ETH3D} \\
    \cmidrule(lr){4-4} \cmidrule(lr){5-5}
    & & &  \multicolumn{1}{c}{AbsRel↓} &  \multicolumn{1}{c}{AbsRel↓} \\
    \midrule      
    \multirow{2}{*}{DA-L} & \multirow{2}{*}{DA-S} &  Base & 0.290 & 0.110\\
     &  & \cellcolor[rgb]{ .886,  .937,  .855} Ours & \cellcolor[rgb]{ .886,  .937,  .855} \textbf{0.262} \green{-9.6} & \cellcolor[rgb]{ .886,  .937,  .855} \textbf{0.098} \green{-10.9} \\
    \midrule
    \multirow{2}{*}{DA-L} & \multirow{2}{*}{Midas-L} &  Base  & 0.313 & 0.147\\
     &  & \cellcolor[rgb]{ .886,  .937,  .855}Ours & \cellcolor[rgb]{ .886,  .937,  .855}\textbf{0.295}\green{-5.7} & \cellcolor[rgb]{ .886,  .937,  .855}\textbf{0.126}\green{-14.3} \\
    \midrule
    \multirow{2}{*}{Midas-L} & \multirow{2}{*}{Midas-S} & Base  & 0.303 & 0.150 \\
     &  & \cellcolor[rgb]{ .886,  .937,  .855}Ours & \cellcolor[rgb]{ .886,  .937,  .855}\textbf{0.272} \green{-10.2}& \cellcolor[rgb]{ .886,  .937,  .855}\textbf{0.120}\green{-20.0}\\
    \bottomrule
    \end{tabular}
    }

\end{table}

\noindent \textbf{Ablation Study of Cross-Context Distillation.}
To further validate the effectiveness of our distillation framework, we conduct ablation studies by removing Shared-Context Distillation and Local-Global Distillation in Table~\ref{tab:performance_comparison}. Without both components, the model degrades to a conventional distillation setup, resulting in significantly lower performance. Introducing Shared-Context Distillation with Hybrid Normalization notably improves accuracy, highlighting the benefits of a better normalization strategy with consistent context supervision. When using only Local-Global Distillation, the model still performs well, showing the effectiveness of region-wise depth refinement even without global context information. Combining both strategies yields the best results, confirming that both components contribute significantly to improving the student model’s ability to utilize pseudo-labels, demonstrating the robustness of our approach.

\noindent \textbf{Cross-Architecture Distillation.} 
To highlight the limitations of previous 
state-of-the-art distillation approaches employing global normalization, we compare their performance against the Hybrid Normalization strategy, which we utilize in our distillation framework, across diverse model architectures. To demonstrate the generalizability of our approach, we conduct cross-architecture distillation experiments on both the state-of-the-art DepthAnything~\cite{depth_anything_v2} and the classic MiDaS~\cite{ranftl2020midas} architecture. Experiments are conducted using MiDaS~\cite{ranftl2020midas} and DepthAnything~\cite{depth_anything_v2} in four configurations (DA-L, MiDaS-L, DA-S, MiDaS-S), as shown in Table~\ref{tab:cross_architecture_distillation}. Our method consistently outperforms previous global normalization-based distillation on both the DIODE~\cite{vasiljevic2019diode} and ETH3D~\cite{schoeps2017eth3d} datasets. These results demonstrate superior performance both within and across architectures, underscoring the limitations of global normalization in pseudo-label distillation.

\begin{table}[t]
    \caption{
    \textbf{Effect of Assistant-Guided Distillation.} 
    Bold values indicate the best performance. Our method integrates a primary teacher, DepthAnything v2 (denoted as ‘D’), with a diffusion-based assistant, GenPercept (denoted as ‘G’), leveraging their complementary strengths to produce higher-quality pseudo-labels. The student model, trained under this assistant-guided distillation framework, consistently achieves better accuracy than when distilled from the DAv2-Large teacher alone.
    }
    \label{tab:multi-teacher}
    \resizebox{0.5\textwidth}{!}{ 
    \begin{tabular}{ccll}
    \toprule
    \multirow{2}{*}{\makecell{Method}} & \multirow{2}{*}{\makecell{Assistant-Guided\\ Strategy}} & \multicolumn{1}{c}{ETH3D} & \multicolumn{1}{c}{DIODE} \\
    & & \multicolumn{1}{c}{AbsRel↓} & \multicolumn{1}{c}{AbsRel↓} \\
    \midrule
    DepthAnything v2 & w/o & 0.131 & 0.262 \\
    Genpercept(Disparity) & w/o & 0.096  & \textbf{0.226} \\
    D + G & Avg. & 0.228 & 0.371  \\
     D + G & Select. & \textbf{0.054}  & 0.258 \\
    \bottomrule
    \end{tabular}
    }
        
\end{table}

\noindent \textbf{Effect of Assistant-Guided Distillation.}
To validate the effectiveness of our proposedassistant-guided distillation strategy, we design a comparative experiment that introduces an additional assistant model to the conventional teacher-student distillation framework. Specifically, we adopt DepthAnything v2 as the primary teacher and GenPercept—a diffusion-based model—as the assistant. The student model shares the same architecture as DAv2-Large and is initialized with its pre-trained weights.
This setup allows us to investigate whether supervision from two diverse architectures—trained under different paradigms—can offer complementary guidance that enhances both generalization and depth estimation performance.
To explore the most effective way to combine pseudo-labels from the primary teacher and the assistant, we compare two assistant-guided strategies: (1) a weighted averaging approach (Avg.), which assigns greater weight to pixels where the two teachers exhibit high agreement, and (2) a selection-based strategy (Select.), which probabilistically samples the supervision signal from either teacher.
While the averaging strategy attempts to leverage consistency between teachers, it often performs poorly due to conflicting pseudo-labels, where averaging can amplify errors. In contrast, the selection-based strategy allows the student to selectively absorb the strengths of each teacher, avoiding error reinforcement.
As shown in Table~\ref{tab:multi-teacher}, the Select. strategy significantly outperforms both individual teachers and the averaging method on the ETH3D benchmark, demonstrating the effectiveness ofassistant-guided distillation in delivering robust and diverse supervision.

\begin{table*}[t]
    
    \centering
\renewcommand{\arraystretch}{1.1}
\caption{
    \textbf{Quantitative comparison with other affine-invariant depth estimators on several zero-shot benchmarks.} The \textbf{bold} values indicate the best performance, and \underline{underscored} represent the second-best results.}
    \label{benchmark}
    \resizebox{\linewidth}{!}{
\begin{tabular}{
    @{}l 
    c@{\hspace{0.5em}}c @{}p{1em}@{} 
    c@{\hspace{0.5em}}c @{}p{1em}@{} 
    c@{\hspace{0.5em}}c @{}p{1em}@{} 
    c@{\hspace{0.5em}}c @{}p{1em}@{} 
    c@{\hspace{0.5em}}c @{}p{1em}@{}
}
    \toprule
    \multirow{2}{*}{Method} &
    \multicolumn{2}{c}{NYUv2} & &
    \multicolumn{2}{c}{KITTI} & &
    \multicolumn{2}{c}{DIODE} & &
    \multicolumn{2}{c}{ScanNet} & &
    \multicolumn{2}{c}{ETH3D} \\
    
    & AbsRel↓ & $\delta$1↑ & &
    AbsRel↓ & $\delta$1↑ & &
    AbsRel↓ & $\delta$1↑ & &
    AbsRel↓ & $\delta$1↑ & &
    AbsRel↓ & $\delta$1↑ \\

    \midrule

    DiverseDepth~\cite{yin2020diversedepth} & 
    0.117 & 0.875 & & 
    0.190 & 0.704 & &
    0.376 & 0.631 & &
    0.108 & 0.882 & &
    0.228 & 0.694 \\

    MiDaS~\cite{ranftl2020midas} & 
    0.111 & 0.885 & & 
    0.236 & 0.630 & &
    0.332 & 0.715 & &
    0.111 & 0.886 & &
    0.184 & 0.752 \\

    LeReS~\cite{wei2021leres} & 
    0.090 & 0.916 & & 
    0.149 & 0.784 & &
    0.271 & 0.766 & &
    0.095 & 0.912 & &
    0.171 & 0.777 \\

    Omnidata~\cite{eftekhar2021omnidata} & 
    0.074 & 0.945 & & 
    0.149 & 0.835 & &
    0.339 & 0.742 & &
    0.077 & 0.935 & &
    0.166 & 0.778 \\

    HDN~\cite{zhang2022hdn} & 
    0.069 & 0.948 & & 
    0.115 & 0.867 & &
    0.246 & \underline{0.780} & &
    0.080 & 0.939 & &
    0.121 & 0.833 \\

    DPT~\cite{ranftl2021dpt} & 
    0.098 & 0.903 & & 
    0.100 & 0.901 & &
    \underline{0.182} & 0.758 & &
    0.078 & 0.938 & &
    0.078 & 0.946 \\


    DepthAnything v2~\cite{yang2024depthanything} & 
    \underline{0.045} & 0.979 & & 
    0.074 & 0.946 & &
    0.262 & 0.754 & &
    \textbf{0.042} & \underline{0.978} & &
    0.131 & 0.865 \\

    GenPercept~\cite{xu2024diffusion} &
    0.058 & 0.969 & &
    0.080 & 0.934 & &
    0.226 & 0.741 & &
    0.063 & 0.960 & &
    0.096 & 0.959 \\

    Marigold~\cite{ke2024marigold} & 
    0.055 & 0.961 & & 
    0.099 & 0.916 & &
    0.308 & 0.773 & &
    0.064 & 0.951 & &
    0.065 & 0.960 \\
    
    MiDaS v3.1
    ~\cite{birkl2023midasv31model} & 
    - & 0.980 & & 
    - & \underline{0.949} & &
    - & - & &
    - & - & &
    0.061 & 0.968 \\

    \midrule
    Ours$^\dagger$ & 
    0.046 & \textbf{0.985} & & 
    \textbf{0.063} & \textbf{0.972} & & 
    \textbf{0.142} & \textbf{0.788} & & 
    0.049 & \textbf{0.980} & & 
    \underline{0.057} & \underline{0.976} \\

    Ours$^*$ &
    \textbf{0.043} & \underline{0.981} & &
    \underline{0.070} & \underline{0.949} & &
    0.233 & 0.753 & &
    \underline{0.043} & \textbf{0.980} & &
      \textbf{0.054} & \textbf{0.981}\\
    \bottomrule
\end{tabular}
    }
    \\
    \begin{minipage}{0.96\linewidth}
        \scriptsize
        \begin{itemize}
        \item[$^\dagger$] Cross-Context distillation applied to MiDaS v3.1, using a pre-trained MiDaS v3.1 model as the teacher.
        \item[$^*$] Cross-Context distillation applied to DepthAnythingv2-Large, using a pre-trained DAv2 model as the teacher.
        \end{itemize}
    \end{minipage}
    \vspace{-0.1in}
    
\end{table*}


\subsection{Comparison with State-of-the-Art}
\label{sec:analysis_zero_shot}

\noindent \textbf{Quantitative Analysis.}
As shown in Table~\ref{benchmark}, our method achieves state-of-the-art performance across a diverse range of zero-shot depth estimation benchmarks. These include both structured indoor scenes (e.g., NYUv2~\cite{silberman2012indoor}, ScanNet~\cite{dai2017scannet}) and challenging outdoor environments (e.g., KITTI~\cite{geiger2012we}, DIODE~\cite{vasiljevic2019diode}, ETH3D~\cite{schoeps2017eth3d}), demonstrating strong generalization across domains with varying scene structures, lighting conditions, and depth statistics.
To further validate the effectiveness and scalability of our distillation framework, we conduct evaluations on two representative model architectures: DepthAnythingv2, a recent state-of-the-art model based on DINOv2, and MiDaS, a classic and widely adopted encoder-decoder framework. For each setup, the student model is initialized with the corresponding pre-trained encoder and distilled using pseudo-labels generated by the teacher model. 
Our approach yields consistent improvements over both teacher models across all benchmarks, highlighting its effectiveness in learning from pseudo-labels. Notably, it establishes new state-of-the-art results in most cases, outperforming existing affine-invariant depth estimators and demonstrating the robustness of our method in both DAv2 and MiDaS settings.
These results confirm that our distillation framework is broadly applicable and can effectively transfer knowledge across model scales and depth distributions, enabling the student model to surpass its teacher in both accuracy and generalization under zero-shot evaluation.

\noindent \textbf{Qualitative analysis.}
We present a qualitative comparison of depth estimations from different models in Fig.~\ref{fig:vis}, including recent state-of-the-art approaches and our student model, which shares the same architecture as DAv2 but is trained using our distillation framework. Compared with DAv2~\cite{depth_anything_v2}, our method clearly preserves finer structural details—particularly in regions highlighted by arrows—thanks to the proposed cross-context distillation strategy and the assistant model's enhanced detail perception.
While diffusion-based MDE methods such as Marigold~\cite{marigold} and GenPercept~\cite{xu2024diffusion} generate visually rich depth maps by leveraging generative priors, they are trained on a limited amount of synthetic data, which hinders their ability to maintain correct relative depth ordering in real-world scenes. This issue arises from the inherent stochasticity and creativity of their generation paradigms, which, although capable of producing accurate depth ordering in certain regions, may introduce inconsistencies in others.
In contrast, our student model effectively balances detail preservation and structural consistency. with shared-context distillation and local-global distillation, it achieves more reliable and robust depth estimations that are both locally detailed and globally consistent.

\section{Conclusion}
In this work, we investigate pseudo-label distillation strategies for MDE. We observe that the commonly used global normalization scheme tends to amplify noise in teacher-generated pseudo-labels, thereby impairing local depth accuracy. To address this issue, we propose Cross-Context Distillation, which combines local refinement with global consistency through a more effective normalization strategy. This enables the model to learn both fine-grained details and high-level structural context. Furthermore, ourassistant-guided distillation framework integrates diffusion-based generative priors as complementary guidance to traditional encoder-decoder networks, achieving state-of-the-art performance across multiple benchmarks.

{\small
\bibliographystyle{ieee_fullname}
\bibliography{egpaper_final}
}


\clearpage
\appendix
\section{Appendix}
\section{Dataset Details}
\subsection{Datasets.}  
Our model is trained on \textbf{SA-1B}~\cite{sa1b}, a large-scale dataset comprising high-quality RGB images of diverse indoor and outdoor scenes. These high-fidelity images facilitate the generation of more detailed pseudo-labels, enabling robust depth estimation and fine-grained detail learning for real-world applications. SA-1B~\cite{sa1b} is also the dataset employed in DAv2~\cite{depth_anything_v2}.
For evaluation, we use established monocular depth benchmarks:  
\begin{itemize}
    \item \textbf{NYUv2}~\cite{silberman2012indoor}: Indoor depth estimation and semantic segmentation dataset. We evaluate on the official test split of 654 samples.
    \item \textbf{KITTI}~\cite{geiger2012we}:  Autonomous driving dataset with outdoor scenes and high-quality LiDAR ground truth depth. Following prior work, we use the 697-image test split. (Corrected number of images based on standard KITTI benchmark)
    \item \textbf{ETH3D}~\cite{schoeps2017eth3d}: High-resolution stereo images for indoor and outdoor depth estimation and 3D reconstruction. We evaluate on all 454 images.
    \item \textbf{ScanNet}~\cite{dai2017scannet}: Large-scale RGB-D dataset for 3D scene reconstruction and semantic segmentation. We use a test split of 1000 samples.
    \item \textbf{DIODE}~\cite{vasiljevic2019diode}: Dense, high-quality depth maps for both indoor and outdoor environments. However, as noted in MoGe~\cite{wang2024moge}, this dataset exhibits artifacts in the depth values of the ground truth near the boundaries of the object.
    
\end{itemize}
For visualization, we also use images from Dav2~\cite{depth_anything_v2}, GenPercept~\cite{xu2024diffusion}, PatchFusion~\cite{patchfusion2023}, Hypersim~\cite{roberts2021hypersim}, Omnidata~\cite{eftekhar2021omnidata}, Depth Pro~\cite{depthpro} and Gen-3.

\subsection{Metrics.}  
We evaluate depth estimation using mean absolute relative error (AbsRel) and $\delta_1$ accuracy. AbsRel is defined as: 
\begin{equation}
AbsRel = \frac{1}{M} \sum_{i=1}^M \frac{|d_i - d_i^*|}{d_i^*}
\end{equation}
where \( d_i \) is the predicted depth, \( d_i^* \) is the ground truth, and \( M \) is the total number of depth values. $\delta_1$ accuracy measures the percentage of pixels where:
\begin{equation}
\delta_1 = \max\left(\frac{d_i}{d_i^*}, \frac{d_i^*}{d_i}\right) < 1.25
\end{equation}
indicating prediction accuracy within a specific tolerance. Following Metric3D~\cite{ranftl2020midas,yin2023metric3d, marigold}, we align predictions with ground truth in scale and shift before evaluation.  


\section{More Experiments}
\subsection{Implementation Details.}
For visualization, our model uses Dav2 as the student model and is fine-tuned with Dav2 parameters as the pre-trained weights. Since our training iterations and dataset size are relatively small, leveraging the strong prior knowledge of Dav2 allows us to achieve significant visual improvements quickly.
Regarding Table~\ref{benchmark}, the model is fine-tuned with Dav2 but only uses the backbone parameters from Dav2. Other components, such as the DPT head, are initialized randomly. We found that training entirely with Dav2's pre-trained parameters does not directly demonstrate the effectiveness of our method. By retaining only the encoder and training the decoder from scratch, the accuracy clearly shows the improvement in pseudo-label utilization due to our normalization strategy, as well as the effectiveness of our cross-context distillation approach.

\subsection{Effect of Data Scaling.}
\label{sec:data_scaling_effect}
\begin{figure}[t]
    \vspace{-1em}
    \centering
    \begin{minipage}{0.4\textwidth}  
        \centering
        \includegraphics[width=\textwidth]{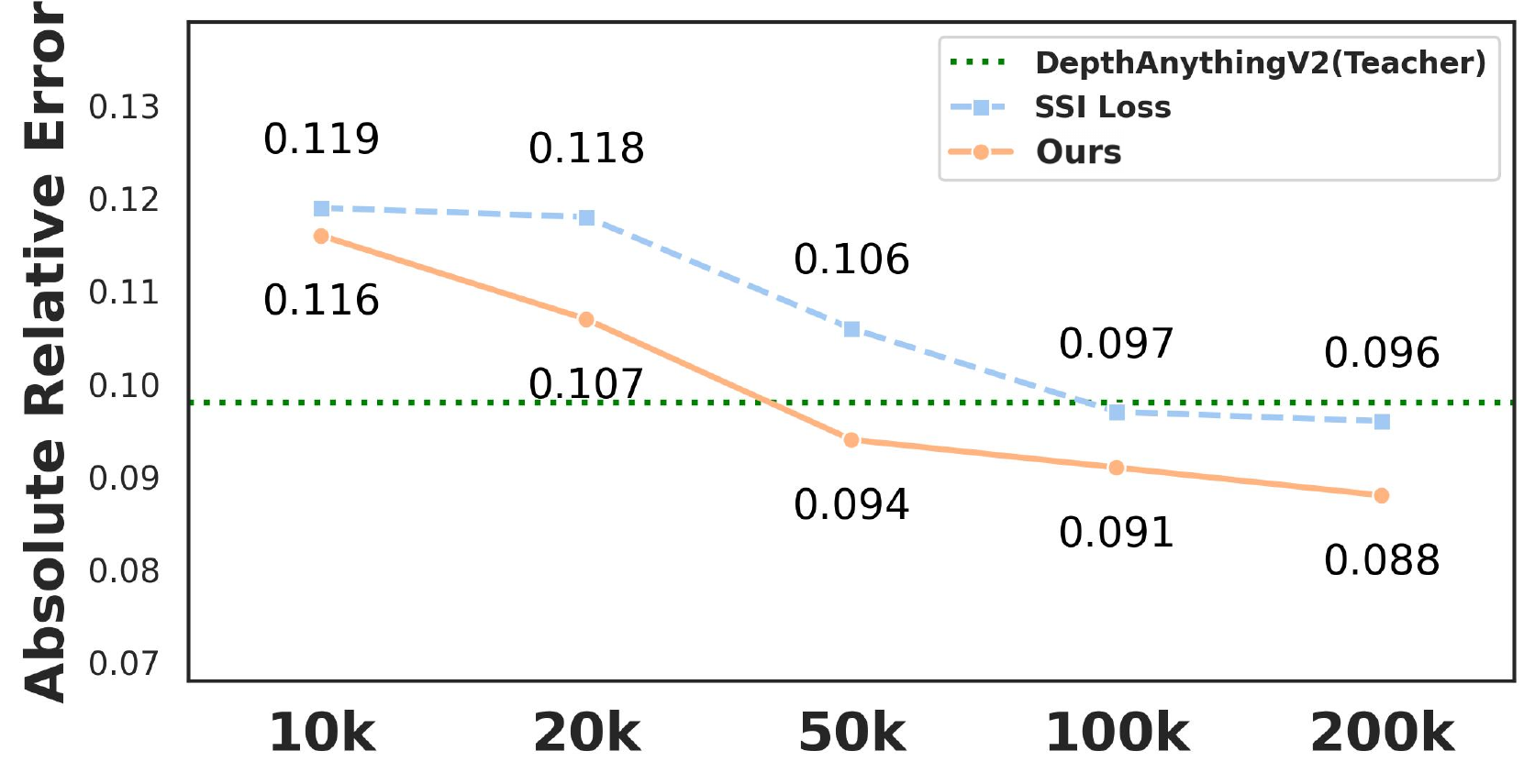}
        \vspace{-2em}
        \caption{\textbf{Comparison of Data Scaling 
        .} Performance comparison of our model with SSI Loss as the dataset size increases, measured by the average AbsRel. The results indicate that our method consistently outperforms the baseline method.
        }
        \label{fig:data_scaling_comparison}
    \end{minipage}
\end{figure}

To investigate the impact of dataset size on model performance, we conducted experiments with progressively larger training sets and compared our method against the SSI Loss baseline across five popular benchmarks. The results are averaged over these benchmarks. As shown in Fig.~\ref{fig:data_scaling_comparison}, we report the Absolute Relative Error (AbsRel) as the dataset size increases from 10K to 200K images. Our distillation pipeline consistently outperforms the traditional SSI-based global normalization approach across all dataset sizes. Notably, the performance gap between our method and the baseline widens as more training data is introduced. Moreover, our approach enables the student model to surpass the teacher’s performance using significantly less training data, highlighting its data efficiency.

\subsection{Distilling Generative Models vs. DepthAnythingv2.}  
\begin{figure*}[h]  
    \centering  
    \includegraphics[width=1\textwidth]{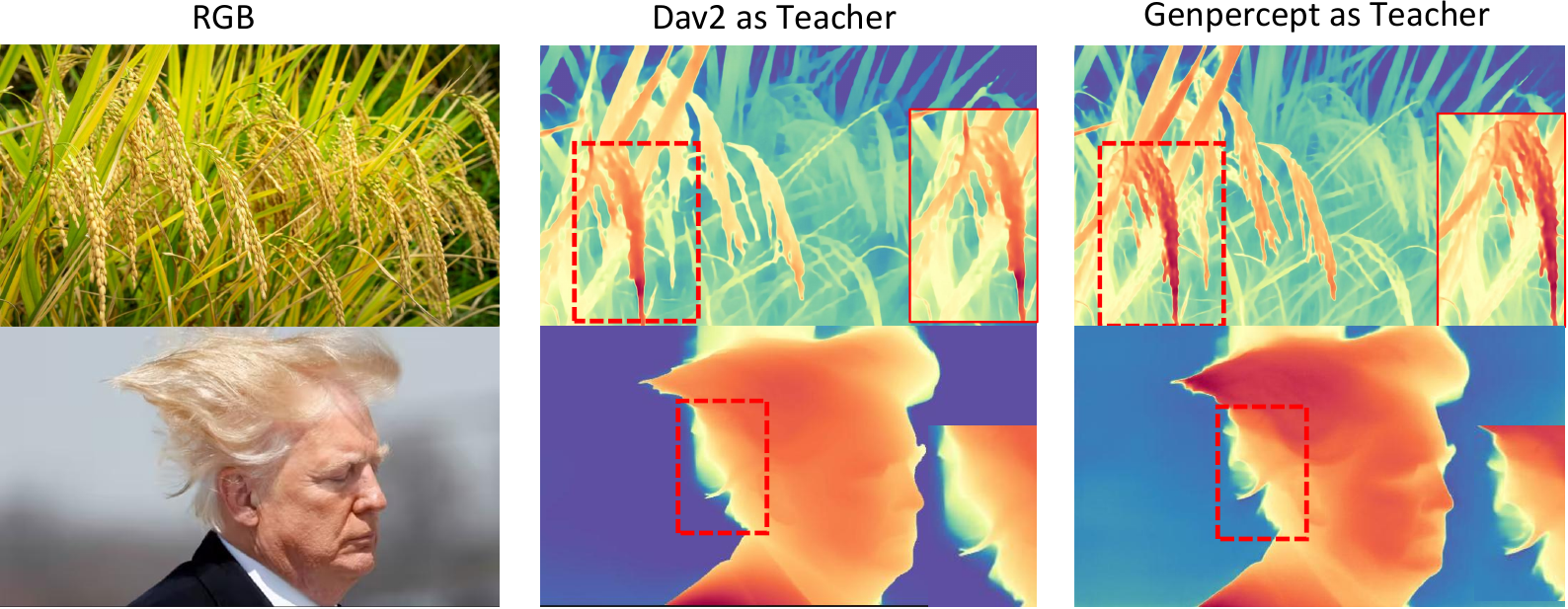}  
    \caption{\textbf{Distilled Generative Models}: Instead of just distilling classical depth models, we also apply distillation to diffusion-based generative models, aiming for the student model to learn the rich details inherent in these models, which are often not fully reflected in standard accuracy metrics.
    }
    \label{fig:distill_with_generation}
\end{figure*}

Beyond distilling encoder-decoder depth models, we extend our approach to generative models, specifically GenPercept~\cite{xu2024diffusion}, aiming to transfer their superior detail preservation to a more efficient student model. While diffusion-based depth estimators achieve fine-grained depth reconstruction, their high computational cost limits practical applications. We investigate whether their depth estimation capability can be effectively distilled into a lightweight DPT-based model. Experimental results in Fig.~\ref{fig:distill_with_generation} show that compared to using DepthAnythingv2 as the teacher, distilling from a diffusion-based model yields a student model with significantly enhanced fine-detail prediction.

\subsection{Qualitative Comparison with Baseline Distillation.}  
\begin{figure*}[h]  
    \centering  
    \includegraphics[width=1\textwidth]{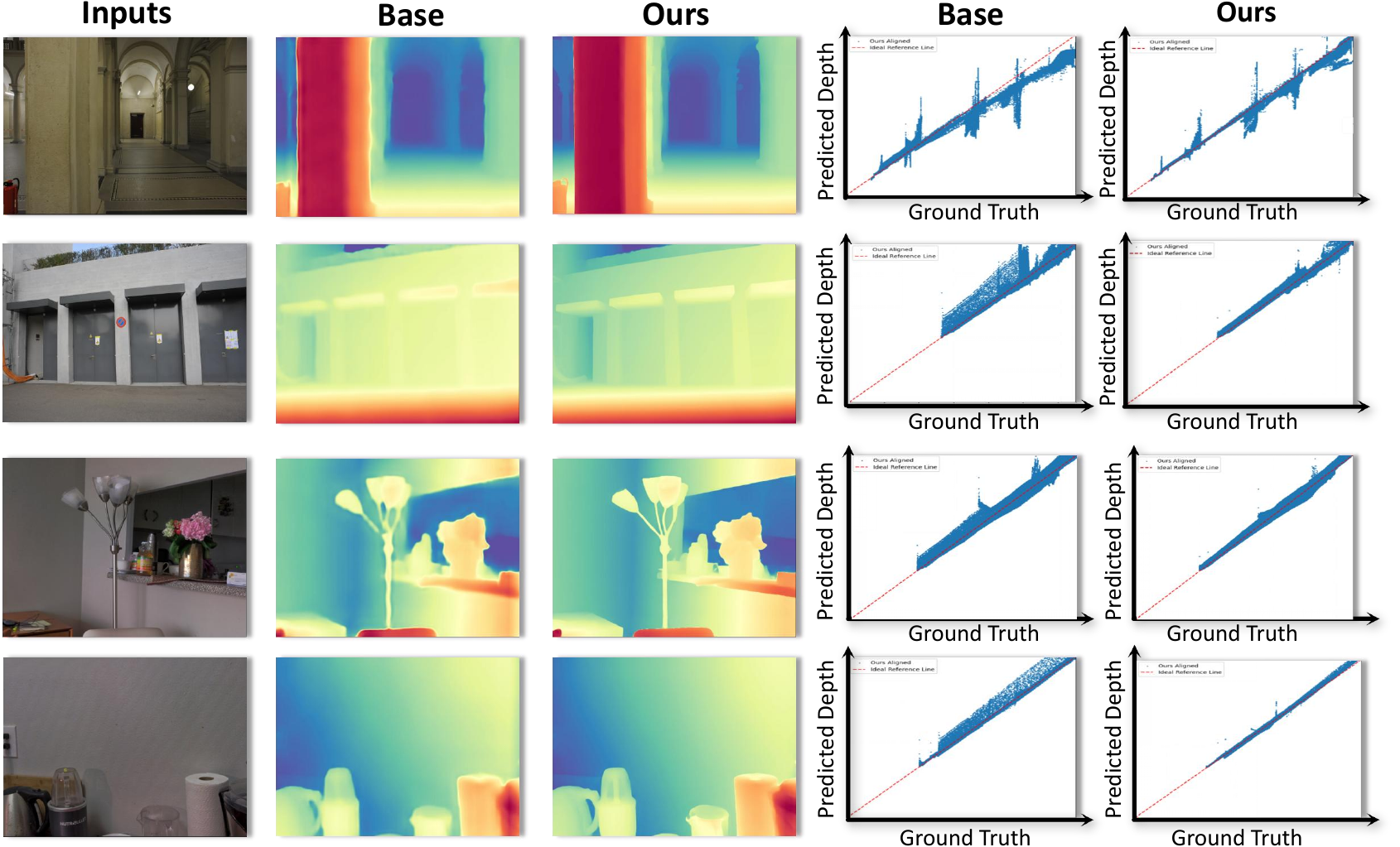}  
    \caption{\textbf{Qualitative Comparison with Baseline Distillation.} We compare our method with the baseline as the previous distillation method, which uses only global normalization. The red diagonal lines represent the ground truth, with results closer to the lines indicating better performance. Our method produces smoother surfaces, sharper edges, and more detailed depth maps.}

    \label{fig:qual_ssi}
\end{figure*}

We present a qualitative comparison between our method and the previous distillation method~\cite{depth_anything_v2}, where the \textbf{Base} model relies solely on global normalization. We analyze the depth map details and the distribution differences between predicted and ground truth depths. The red diagonal lines represent the ground truth, with results closer to these lines indicating better performance. As shown in Fig.~\ref{fig:qual_ssi}, our method produces smoother surfaces, sharper edges, and more detailed depth maps.

\subsection{Additional Results on 3D reconstruction in the Wild.}
\begin{figure*}[h]  
    \centering  
    \includegraphics[width=0.9\textwidth]{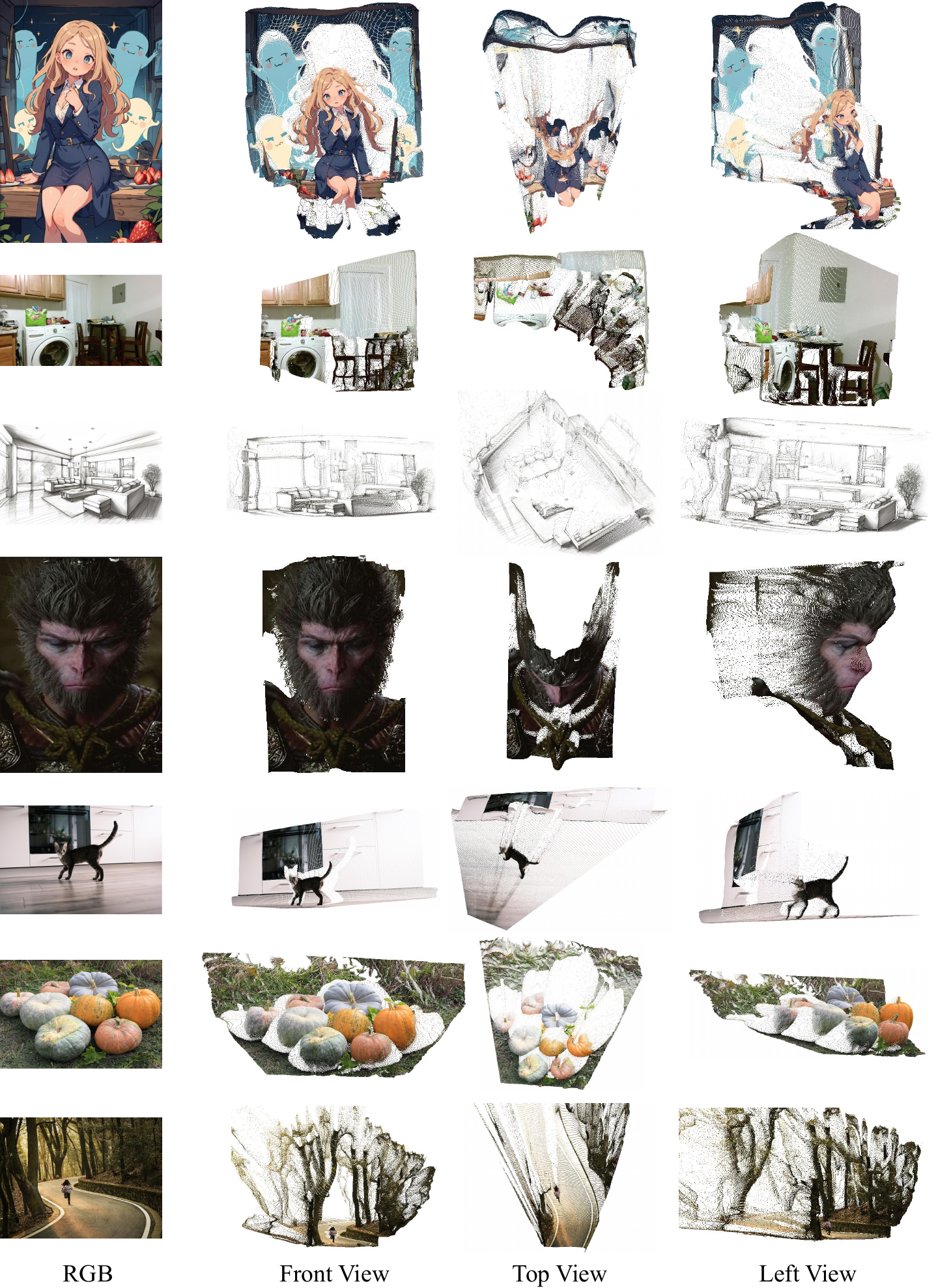}  
    \caption{\textbf{Additional results on 3D reconstruction from in-the-wild RGB images.} We present point clouds generated from our model's predicted depth maps, aligned with geometry-preserving depth from MoGe~\cite{wang2024moge}. These visualizations demonstrate the effectiveness and practical applicability of our model in unconstrained, real-world scenarios.}

    \label{fig:pcd}
\end{figure*}

Benefiting from MoGe's advances in geometry-preserving depth estimation, we align the relative depth predicted by our model with MoGe's outputs. Using the camera parameters estimated by MoGe, we project the aligned depth into 3D space to obtain visualizable point clouds.
As shown in Fig.~\ref{fig:pcd}, these visualizations demonstrate the effectiveness and practical applicability of our model in unconstrained, real-world scenarios. Remarkably, our method performs well even on stylized or synthetic content, such as anime-style images, making it potentially useful for downstream tasks like virtual character modeling. Similarly, our model generates high-quality reconstructions for images from game engines. In real-world photographs—captured by consumer devices such as smartphones—the reconstructed point clouds preserve meaningful geometric structures and fine details. Furthermore, even in abstract scenes like sketches with missing visual cues, our model can infer plausible relative depth and recover a semantically coherent 3D layout of the entire scene.

\subsection{Qualitative Comparison: Additional Results on Depth Estimation in the Wild.} 
\begin{figure*}[h]  
    \centering  
    \includegraphics[width=1\textwidth]{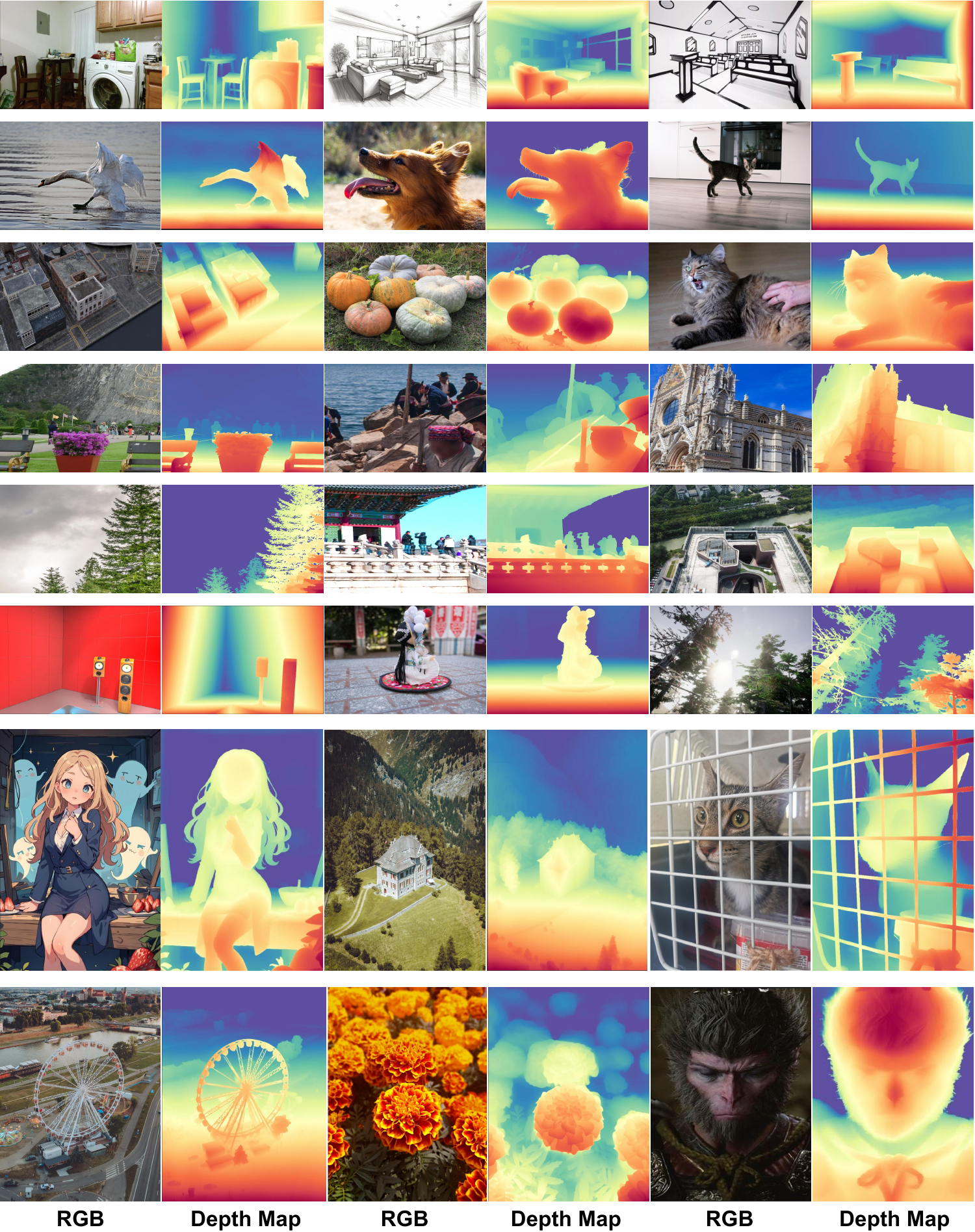}  
    \vspace{-1.5em}
    \caption{\textbf{Additional Results on Depth Estimation in the Wild.} We showcase more depth maps generated by our model on in-the-wild scenes, highlighting its robustness and precision.}

    \label{fig:more_results}
\end{figure*}

As shown in Fig.~\ref{fig:more_results}, our model demonstrates strong generalization and robustness across a wide range of scenarios, including real-world indoor and outdoor environments, stylized virtual content such as anime and game engine renders, and information-sparse inputs like sketches or line drawings. Even in unconventional perspectives such as bird's-eye cityscapes, the model preserves accurate relative depth and structural coherence. These results highlight its ability to deliver detailed and semantically meaningful depth predictions in both natural and synthetic domains, enabling practical applications in 3D reconstruction, content creation, and downstream tasks across real and virtual worlds.

\end{document}